\documentclass[supp]{new_tlp} %% added supp option

\usepackage{supp} %% added

\pubauthor{Zese} %% added, please replace "YourLastName" appropriately
\usepackage{amsmath}
 \usepackage{newfloat}

\usepackage{graphicx}
\usepackage[all]{xy}
\usepackage{epstopdf}
\usepackage{subfigure}
\usepackage{caption}
\usepackage{url}

\newcommand\cI{{\cal I}}

\newcommand\cT{{\cal T}}

\newcommand\cA{{\cal A}}
\newcommand\cK{{\cal K}}

\newcommand\cL{{\cal L}}
\newcommand\cC{{\cal C}}

\newcommand\fluffy{{\mathit{fluffy}}}

\title{Reasoning with Probabilistic Logics}
\author[R. Zese] 
{RICCARDO ZESE \\
Dipartimento di Ingegneria -- University of Ferrara\\
Via Saragat 1, 44122, Ferrara, Italy  \\
\email{riccardo.zese@unife.it}}

\newtheorem{example}{Example}[section]
\newtheorem{definition}{Definition}[section]

\begin{document}

\label{firstpage}

\maketitle

\begin{abstract}
  The interest in the combination of 
  probability with logics for modeling the world has rapidly increased in the last few years. One of the most effective approaches is 
  the Distribution Semantics which was adopted by many logic programming languages
  and in Descripion Logics.
  In this paper, we illustrate the work we have done in this research field by presenting a probabilistic semantics 
  for description logics and reasoning and learning algorithms. In particular, we present in detail the system 
  TRILL$^P$, which computes the probability of queries w.r.t. probabilistic knowledge bases, which has been implemented in Prolog.
  {\bf Note:} An extended abstract / full version of a paper accepted to be presented at the Doctoral Consortium of the 30th International Conference on Logic Programming (ICLP 2014), July 19-22, Vienna, Austria
\end{abstract}


\section{Introduction}

In the last few years, many researchers tried to combine first-order logic and probability for modeling uncertain domains 
and performing inferece and learning.
In the field of Probabilistic Logic Programming (PLP for short) many proposals have been presented. 
An effective and popular approach is the Distribution Semantics \cite{DBLP:conf/iclp/Sato95}, which underlies many PLP languages such as
 PRISM~\cite{DBLP:conf/iclp/Sato95,DBLP:journals/jair/SatoK01},
 Independent Choice Logic  \cite{Poo97-ArtInt-IJ}, Logic Programs with Annotated Disjunctions \cite{VenVer04-ICLP04-IC} and ProbLog \cite{DBLP:conf/ijcai/RaedtKT07}.
 Along this line, many reserchers proposed to combine probability theory with Description Logics (DLs for short) \cite{DBLP:journals/ws/LukasiewiczS08,DBLP:conf/rweb/Straccia08}.
DLs are at the basis of the Web Ontology Language (OWL for short), a family of knowledge representation formalisms used for modeling information
 of the Semantic Web
%  , that aims at making information automatically manageable
% by machines 
\cite{sw}.
 In \cite{RigBelLamZese12-URSW12} we presented DISPONTE, 
a probabilistic semantics for DLs based on the distribution semantics that allows probabilistic assertional and terminological knowledge.

In order to allow inference over the information in the Semantic Web, many efficient DL reasoners, such as  Pellet \cite{DBLP:journals/ws/SirinPGKK07}, 
RacerPro \cite{DBLP:dblp_journals/semweb/HaarslevHMW12} and HermiT \cite{DBLP:dblp_conf/owled/ShearerMH08}, have been developed. 
Despite the availability of many DL reasoners, the number of probabilistic reasoners is quite small. In \cite{RigBelLamZese13-RR13} we presented BUNDLE,
a reasoner based on Pellet that extends it by allowing to perform inference on DISPONTE theories. 
Most of the available DL reasoners, included BUNDLE, exploit procedural languages for implementing their reasoning algorithms. Nonetheless, some of them use non-deterministic 
operators for doing inference. We implemented a reasoner, 
called TRILL \cite{DBLP:conf/cilc/ZeseBLR13}, that exploits Prolog for managing the non-determinism. 
Then, we developed a new version of TRILL, called TRILL$^P$ and we added in both versions the ability to manage DISPONTE knowledge bases (KBs for short)
and computing the probability of a query given a probabilistic KB under the DISPONTE semantics.

Since a problem of probabilistic KBs is that the parameters are difficult to define, in \cite{DBLP:conf/rr/RiguzziBLZ13a} we presented 
EDGE that learns the parameters of a DISPONTE KB from the information available in the domain.
Moreover, we are currently  working on the extension of EDGE in order also to learn the structure of the probabilistic KB togheter with the parameters.
% Finally, we briefly present LP$^2$, an acronim for ``Lifted Probabilistic Logic Programming'', a new approach under study for performing lifted inference for logic programming under the distribution semantics 
% and exploits lifted variable elimination for computing the probability of given queries.

In the field of PLP, we are working at improving existing algorithms. We have considered lifted inference that allows to perform inference in a time 
that is polynomial in the variables' domain size. We applied lifted variable elimination, and GC-FOVE \cite{DBLP:journals/jair/TaghipourFDB13} in particular, to 
PLP and developed the algorithm LP$^2$.

The paper is organised as follows. Section \ref{DL} briefly introduces $\mathcal{ALC}$, while Section \ref{disp} presents the DISPONTE semantics.
Section \ref{problem} defines the problem of finding explanations for a probabilistic query w.r.t. a given probabilistic KB. Section \ref{trill} presents TRILL and TRILL$^P$ and Section \ref{related}
 discusses related work. Section \ref{exp} shows experiments and section \ref{issues-achi} discusses our achievements and future plans. Finally, Section~\ref{conc} concludes the paper.

\section{Description Logics}
\label{DL}

DLs are knowledge representation formalisms that are at the basis of the Semantic Web \cite{DBLP:conf/dlog/2003handbook,dlchap} and are used for modeling ontologies.
They are represented using a syntax based on concepts, basically sets of individuals of the domain, and roles, sets of pairs of individuals
of the domain. In this section, we recall the expressive description logic $\mathcal{ALC}$ \cite{DBLP:journals/ai/Schmidt-SchaussS91}. We refer to 
\cite{DBLP:journals/ws/LukasiewiczS08} for a detailed description of $\mathcal{SHOIN}(\mathbf{D})$ DL, that is at the basis of OWL DL.

%%ALC
Let $\mathbf{A}$, $\mathbf{R}$ and $\mathbf{I}$ be sets of \emph{atomic concepts}, \emph{roles} and \emph{individuals}.
A \emph{role} is an atomic role $R \in \mathbf{R}$. 
\emph{Concepts} are defined by induction as follows. Each $C \in \mathbf{A}$, $\bot$ and $\top$
are concepts.
If $C$, $C_1$ and $C_2$ are concepts and $R \in \mathbf{R}$, then $(C_1\sqcap C_2)$, $(C_1\sqcup C_2 )$, $\neg C$,
$\exists R.C$, and $\forall R.C$ are concepts. 
Let $C$, $D$ be concepts,  $R \in \mathbf{R}$ and $a, b \in \mathbf{I}$. 
An \emph{ABox} $\cA$ is a finite set of \textit{concept membership axioms} $a : C$ and \textit{role membership
axioms} $(a, b) : R$, while 
a \emph{TBox} $\cT$ is a finite set of \textit{concept inclusion axioms} $C\sqsubseteq D$. $C \equiv D$ abbreviates $C \sqsubseteq D$ and $D\sqsubseteq  C$.

A \emph{knowledge base} $\cK = (\cT, \cA)$ consists of a TBox $\cT$ and an ABox $\cA$.
A KB $\cK$ is assigned a semantics in terms of set-theoretic interpretations $\cI = (\Delta^\cI , \cdot^\cI )$, where $\Delta^\cI$ is a non-empty \textit{domain} and $\cdot^\cI$ is the \textit{interpretation function} that assigns  an element in $\Delta ^\cI$ to each $a \in \mathbf{I}$, a subset of $\Delta^\cI$ to each $C \in \mathbf{A}$ and a subset of $\Delta^\cI \times \Delta^\cI$ to each $R \in \mathbf{R}$.

A query $Q$ over a KB $\cK$ is an axiom for which we want to test the entailment from the knowledge base, written $\cK \models Q$. 
The entailment test may be reduced to checking the unsatisfiability of a concept in the knowledge base, i.e., the emptiness of the concept.
 For example, the entailment of the axiom $C\sqsubseteq D$ may be tested by checking the satisfiability of the concept $C \sqcap \neg D$.

\section{The DISPONTE Semantics}
\label{disp}
DISPONTE \cite{RigBelLamZese12-URSW12} applies the distribution semantics \cite{DBLP:conf/iclp/Sato95} of probabilistic logic programming to DLs. 
A program following this semantics defines a probability distribution over normal logic programs
called \emph{worlds}. Then the distribution is extended to queries and the probability of a query is obtained by marginalizing the joint distribution of the query and the programs.

In DISPONTE, a \emph{probabilistic knowledge base} $\cK$ is a set of \emph{certain axioms} or \emph{probabilistic axioms} in which each axiom is independent evidence.
%****************************************%
Certain axioms take the form of regular DL axioms while probabilistic axioms are
% \begin{equation}
% p::E\label{pax}
% \end{equation}
$p::E$
where $p$ is a real number in $[0,1]$ and $E$ is a DL axiom. 

The idea of DISPONTE is to associate independent Boolean random variables to the probabilistic axioms. 
To obtain a \emph{world}, we include every formula obtained from a certain axiom. 
For each probabilistic axiom, we decide whether to include it or not in $w$.
A world therefore is a non probabilistic KB that can be assigned a semantics in the usual way.
A query is entailed by a world if it is true in every  model of the world.

The probability $p$ can be interpreted as an \emph{epistemic probability}, i.e., as the degree of our belief in axiom $E$. 
For example, a probabilistic concept membership axiom
$
p::a:C
$
means that we have degree of belief $p$ in $C(a)$.
A probabilistic concept inclusion axiom of the form
$
p::C\sqsubseteq D
$
represents our belief in the truth of $C \sqsubseteq D$ with probability $p$. 

Formally, an \emph{atomic choice} is a couple $(E_i,k)$ where $E_i$ is the $i$th probabilistic axiom  and $k\in \{0,1\}$. 
$k$ indicates whether $E_i$ is chosen to be included in a world ($k$ = 1) or not ($k$ = 0). 
A \emph{composite choice} $\kappa$ is a consistent set of atomic choices, i.e.,  $(E_i,k)\in\kappa, (E_i,m)\in \kappa$ implies $k=m$ (only one decision is taken for each formula). 
The probability of a composite choice $\kappa$  is 
$P(\kappa)=\prod_{(E_i,1)\in \kappa}p_i\prod_{(E_i, 0)\in \kappa} (1-p_i)$, where $p_i$ is the probability associated with axiom $E_i$.
A \emph{selection} $\sigma$ is a total composite choice, i.e., it contains an atomic choice $(E_i,k)$ for every 
probabilistic axiom  of the probabilistic KB. 
A selection $\sigma$ identifies a theory $w_\sigma$ called  a \emph{world} in this way:
$w_\sigma=\cC\cup\{E_i|(E_i,1)\in \sigma\}$ where $\cC$ is the set of certain axioms. Let us indicate with $\mathcal{S}_\cK$ the set of all selections and with $\mathcal{W}_\cK$ the set of all worlds.
The probability of a world $w_\sigma$  is 
$P(w_\sigma)=P(\sigma)=\prod_{(E_i,1)\in \sigma}p_i\prod_{(E_i, 0)\in \sigma} (1-p_i)$.
$P(w_\sigma)$ is a probability distribution over worlds, i.e., $\sum_{w\in \mathcal{W}_\cK}P(w)=1$.

We can now assign probabilities to queries. 
Given a world $w$, the probability of a query $Q$ is defined as $P(Q|w)=1$ if $w\models Q$ and 0 otherwise.
The probability of a query can be defined by marginalizing the joint probability of the query and the worlds, i.e.
% \begin{equation}
% \label{pq}
% P(Q)=\sum_{w\in \mathcal{W}_\cK}P(Q,w)=\sum_{w\in \mathcal{W}_\cK} P(Q|w)p(w)=\sum_{w\in \mathcal{W}_\cK: w\models Q}P(w)
% \end{equation}
$P(Q)=\sum_{w\in \mathcal{W}_\cK}P(Q,w)=\sum_{w\in \mathcal{W}_\cK} P(Q|w)p(w)=\sum_{w\in \mathcal{W}_\cK: w\models Q}P(w)$.

\begin{example}
\label{people+petsxy}
\begin{small}
Consider the following KB, inspired by the \texttt{people+pets} ontology  \cite{ISWC03-tut}:
{\center $0.5\ \ ::\ \ \exists hasAnimal.Pet \sqsubseteq NatureLover\ \ \ \ \ 0.6\ \ ::\ \ Cat\sqsubseteq Pet$\\
$(kevin,tom):hasAnimal\ \ \ \ \ (kevin,\fluffy):hasAnimal\ \ \ \ \ tom: Cat\ \ \ \ \ \fluffy: Cat$\\}
\noindent The KB indicates that the individuals that own an animal which is a pet are nature lovers with a 50\% probability and that $kevin$ has the animals 
$\fluffy$ and $tom$.  Fluffy and $tom$ are cats and cats are pets with probability 60\%.
We associate a Boolean variable to each axiom as follow
%\begin{small}
$F_1 = \exists hasAnimal.Pet \sqsubseteq NatureLover$, $F_2=(kevin,\fluffy):hasAnimal$, $F_3=(kevin,tom):hasAnimal$, $F_4=\fluffy: Cat$, $F_5=tom: Cat$ and $F_6= Cat\sqsubseteq Pet$.
%\end{small}.

The KB has four worlds and the query axiom $Q=kevin:NatureLover$ is true in one of them, the one corresponding to the selection 
$
\{(F_1,1),(F_2,1)\}
$.
%where each pair contains the corresponding axiom and the value of its the selector is $k = 1$.
The probability of the query is $P(Q)=0.5\cdot 0.6=0.3$.
\end{small}
\end{example}

\begin{example}
\label{people+pets_comb}
\begin{small}
Sometimes we have to combine knowledge from multiple, untrusted sources, each one with a different reliability. 
Consider a KB similar to the one of Example \ref{people+petsxy} but where we have a single cat, $\fluffy$.
{\center $\exists hasAnimal.Pet \sqsubseteq NatureLover\ \ \ \ \ (kevin,\fluffy):hasAnimal\ \ \ \ \ Cat\sqsubseteq Pet$\\}

\noindent and there are two sources of information with different reliability that provide the information that $\fluffy$ is a cat. 
On one source the user has a degree of belief of 0.4, i.e., he thinks it is correct with a 40\% probability,  
while on the other source he has a degree of belief 0.3. %, i.e. he thinks it is correct with a 30\% probability. 
The user can reason on this knowledge by adding the following statements to his KB:
{\center$0.4\ \ ::\ \ \fluffy: Cat\ \ \ \ \ 0.3\ \ ::\ \ \fluffy: Cat$\\}
The two statements represent independent evidence on $\fluffy$ being a cat. We associate $F_1$ ($F_2$) to the first (second) probabilistic axiom.

The query axiom $Q=kevin:NatureLover$ is true in 3 out of the 4 worlds, those corresponding to the selections 
$
\{ \{(F_1,1),(F_2,1)\},
\{(F_1,1),(F_2,0)\},
\{(F_1,0),(F_2,1)\}\}
$. 
So 
$P(Q)=0.4\cdot 0.3+0.4\cdot 0.7+ 0.6\cdot 0.3=0.58.$
This is reasonable if the two sources can be considered as independent. In fact,  the probability comes from the  disjunction of two
independent Boolean random variables with probabilities respectively 0.4 and 0.3: 
$
P(Q) = P(X_1\vee X_2) = P(X_1)+P(X_2)-P(X_1\wedge X_2)
 = P(X_1)+P(X_2)-P(X_1)P(X_2)
 = 0.4+0.3-0.4\cdot 0.3=0.58
$
\end{small}
\end{example}

\section{Querying KBs}
\label{problem}
Traditionally, a reasoning algorithm decides  whether an axiom is entailed or not by a KB by refutation: the  axiom $E$ is entailed if $\neg E$ has no model
in the KB.
Besides deciding whether an axiom is entailed by a KB, we want to find also explanations for the axiom.

The problem of finding  explanations for a query
has been investigated by various authors \cite{DBLP:conf/ijcai/SchlobachC03,DBLP:journals/ws/KalyanpurPSH05,Kalyanpurphd,DBLP:conf/semweb/KalyanpurPHS07,extended_tracing}.
 It was called  \emph{axiom pinpointing} in 
\cite{DBLP:conf/ijcai/SchlobachC03}  and considered as a non-standard reasoning service useful for tracing derivations and debugging ontologies. 
In particular, in \cite{DBLP:conf/ijcai/SchlobachC03} the authors define \emph{minimal axiom sets}  (\emph{MinAs} for short).
\begin{definition}[MinA]
 Let $\cK$ be a knowledge base and $Q$ an
axiom that follows from it, i.e., 
$\cK \models Q$. We call a set 
$M\subseteq \cK$ a
\emph{minimal axiom set} or \emph{MinA} for $Q$ in $\cK$ if 
$M \models Q$ and it is minimal
w.r.t. set inclusion.
\end{definition}  
\noindent The problem of enumerating all MinAs is called \textsc{min-a-enum}.
\textsc{All-MinAs($Q,\cK$)} is the set of all MinAs for query $Q$ in knowledge base $\cK$.
Reasoners such as Pellet solve the \textsc{min-a-enum} problem by finding a single MinA using a tableau algorithm and then applying the \emph{hitting set} \cite{DBLP:journals/ai/Reiter87} algorithm for finding all the others.

A \emph{tableau} is a graph where each node represents an
individual $a$ and is labeled with the set of concepts $\cL(a)$ it belongs to. Each
edge $\langle a, b\rangle$ in the graph is labeled with the set of roles to which the couple
$(a, b)$ belongs. Then, a set of  consistency preserving tableau
expansion rules are repeatedly applied until a clash (i.e., a contradiction) is detected or a clash-free
graph is found to which no more rules are applicable. A clash is for example a
couple $(C, a)$ where $C$ and $\neg C$ are present in the label of a node, i.e. ${C, \neg C} \subseteq \cL(a)$.

Some expansion rules are non-deterministic, i.e., they generate
a finite set of tableaux. Thus the algorithm keeps a set of tableaux that is
consistent if there is any tableau in it that is consistent, i.e., that is clash-free.
Each time a clash is detected in a tableau $G$, the algorithm stops applying rules
to $G$. Once every tableau in $T$ contains a clash or no more expansion rules
can be applied to it, the algorithm terminates. If all the tableaux in the final
set $T$ contain a clash, the algorithm returns unsatisfiable as no model can be
found. Otherwise, any one clash-free completion graph in $T$ represents a possible
model for the concept and the algorithm returns satisfiable.
The hitting set algorithm is a black box method: it repeatedly removes an axiom from the KB and then computes again a MinA recording all the different MinAs so found.

\textsc{min-a-enum} is required to answer queries to KBs following the DISPONTE semantics. To
compute the probability of a query, the explanations must be made mutually exclusive, so
that the probability of each individual explanation is computed and summed
with the others. This can be done by exploiting a splitting algorithm as shown in \cite{DBLP:journals/jlp/Poole00}. 
Alternatively, we can assign independent Boolean random variables to the axioms contained in the explanations and defining 
the Disjunctive Normal Form (DNF) Boolean formula $f_K$ which models the set of explanations. Thus
$
f_K(\mathbf{X})=\bigvee_{\kappa\in K}\bigwedge_{(E_i,1)}X_{i}\bigwedge_{(E_i,0)}\overline{X_{i}}
$
where $\mathbf{X}=\{X_{i}|(E_i,k)\in\kappa,\kappa\in K\}$ is the set of Boolean random variables.
We can now translate $f_K$ to a Binary Decision Diagram (BDD), from which we can compute the probability of the query with a dynamic programming algorithm that is linear in the size of the BDD.
% % A BDD for a  function of Boolean variables is   
% % a rooted graph that has one level for each Boolean variable. 
% % A node $n$ in a BDD has two children: one corresponding to the 1 value of the variable associated with the level of $n$, indicated with $child_1(n)$, and one corresponding the 0 value of the variable, indicated with $child_0(n)$. When drawing BDDs, the 0\--branch - the one going to  $child_0(n)$ -  is distinguished from the 1-branch by drawing it with a dashed line.
% % The leaves store either 0 or 1.
% % A BDD for the function
% % $
% % f(\mathbf{X})=(X_{1}\wedge X_{3})\vee (X_{2}\wedge X_{3})
% % $
% % is shown in Figure \ref{dd}.
% % %The BDD for Example \ref{epidemy} is shown in Figure \ref{dd}.
% % \begin{figure}
% % 	$$
% % \xymatrix@=2mm
% % { X_{1} & &*=<25pt,10pt>[F-:<3pt>]{n_1}
% % \ar@/_/@{-}[ldd] \ar@/^/@{--}[dr]\\ 
% % X_{2}  & & &*=<25pt,10pt>[F-:<3pt>]{n_2} 
% % \ar@/_/@{-}[dll]\ar@{--}[dd] 
% % \\
% % X_{3}& *=<25pt,10pt>[F-:<3pt>]{n_3}
% % \ar@{-}[d] \ar@/^/@{--}[drr]  \\
% % &*=<25pt,10pt>[F]{1} &&*=<25pt,10pt>[F]{0}}
% % $$	
% % \caption{BDD for function  $f(\mathbf{X})$.}
% % \label{dd}
% % \end{figure}
% % 

\section{The algorithms TRILL and TRILL$^P$}
\label{trill}

% \section{Goal and Status of the Research}
% Our main aim is to create a complete framework that allows one to manage big amount of (probabilistic) information in an automatically way.
% The DISPONTE semantics is a probabilistic DL that permits the use of epistemic probability. Moreover, is under study its expansion in order to insert 
%  statistical probability that exploits instantiated explanations for computing the probability of a given query.
% Additionally, we have implemented two DL reasoners, BUNDLE and TRILL, and a learning algorithm, EDGE.
% 
% \subsection{BUNDLE}
% BUNDLE computes the probability of queries from a probabilistic knowledge bases that follow the DISPONTE semantics. It is based on Pellet \cite{DBLP:journals/ws/SirinPGKK07} and is completely written in Java. 
% It modifies Pellet's behavior in order to solve the \textsc{inst-min-a-enum} problem.
% BUNDLE first find an InstMinA and then exploits the hitting set algorithm for computing all the other explanationsBoth the tableau and the hitting set algorithms are modified version
% of those implemented in Pellet. Thereafter, it translates the set of instantiated explanations into a Binary Decision Diagram (BDD) that is
% a data structure that allows the computation of the probability with a dynamic programming algorithm in polynomial time in the size of the
% diagram. Finally, it computes the probability of the query on the built BDD.
% 
% \subsection{TRILL}
TRILL \cite{DBLP:conf/cilc/ZeseBLR13} implements the tableau algorithm using Prolog. In this way, we do not have to implement a search strategy, such as the hitting set algorithm, 
because the management of the non-determinism is demanded to Prolog.
TRILL %exploits the Thea2 library \cite{DBLP:conf/semweb/VassiliadisWM08} that converts OWL DL ontologies into Prolog facts. 
takes as input an OWL DL ontology translated into Prolog facts by using the Thea2 library \cite{DBLP:conf/semweb/VassiliadisWM08}.
For example, a subclass axiom $Cat \sqsubseteq Pet$ is translated into \texttt{subClass('Cat','Pet')}
while for more complex axioms, Thea2 uses Prolog lists, so the axiom $NatureLover \equiv PetOwner \sqcup GardenOwner$ is translated into \linebreak
\texttt{equivalentClasses(['NatureLover',unionOf(['PetOwner','GardenOwner'])])}.

TRILL builds a tableau following the tableau algorithm. 
The non-deterministic rules are treated differently from the deterministic ones. 
While the latter ones are implemented by predicates that take as input a tableau and return a single tableau,
the former ones are implemented by predicates that take as input a tableau but return a list of tableaux from which one is non-deterministically chosen. 
The computation of \textsc{All-MinAs($Q,\cK$)} is performed by simply calling \texttt{findall/3} over the tableau predicate.

% \subsection{EDGE}
% EDGE is a algorithm that can learn parameters for probabilistic DL ontologies that follows the DISPONTE semantics. It takes as input a DL theory and a number of examples that represent the queries.
% Usually, the queries are concept assertion divided into two different sets: positive examples that contains information that we regard as true and for which we would like to get
% high probability and negative examples that represents information that we know are false and for which we would like to get a low probability.
% EDGE first computes, for each example, the BDD encoding its explanations, then it enter in an \emph{Expectation-Maximization} (EM) algorithm, in which the functions 
% Expactation and Maximization are repeatedly applied until the log-likelihood of the examples reaches a local maximum. 
%  

A new version of TRILL, called TRILL$^P$, resolves the axiom pinpointing problem by computing 
a \emph{pinpointing formula}  \cite{DBLP:journals/jar/BaaderP10,DBLP:journals/logcom/BaaderP10} instead of a set of MinAs. 
To define the pinpointing formula we first have to associate a Boolean variable to each axiom of the KB $\cK$. 
The pinpointing formula is a monotone Boolean formula on these variables. This formula compactly encodes the set of all MinAs.
 Let assume that each axiom $E$ of a KB $\cK$ is associated with the propositional variable $var(E)$. The set of all propositional variables is indicated with $var(\cK )$.
 A valuation $\nu$ of a monotone Boolean formula is the set of propositional variables that are true. For a valuation $\nu \subseteq var(\cK)$, let $\cK_{\nu} := \{t \in \cK |var(t)\in\nu\}$.
 \\
 \begin{definition}[Pinpointing formula]
 Given a query $Q$ and a KB $\cK$, a monotone Boolean formula $\phi$ over $var(\cK)$ is called a \emph{pinpointing formula} for $Q$ if for every valuation $\nu \subset var(\cK)$ it holds that $\cK_{\nu} \models Q$ if 
 $\nu$ satisfies $\phi$.
  
 \end{definition}
 In Lemma 2.4 of \cite{DBLP:journals/logcom/BaaderP10}, the authors proved that we can obtain all MinAs from a pinpointing formula  by transforming the formula into DNF and removing disjuncts implying other disjuncts.
%  Thus, the BDD that models the pinpointing formula corresponds to the BDD built following the set of all MinAs.
%The example below illustrates axiom pinpointing and the pinpointing formula.
% The pinpointing formula is a boolean formula built using the variables and the conjunction and disjunction connectives.
From this formula, the construction of BDD can be performed as for MinAs.
For formal definitions see \cite{DBLP:journals/jar/BaaderP10,DBLP:journals/logcom/BaaderP10}.
%The example below illustrates axiom pinpointing and the pinpointing formula.
\begin{example}[Pinpointing formula]
\label{people+pets}
\begin{small}
Consider the KB of Example \ref{people+petsxy} with the same association between Boolean variables and axioms.
Let $Q=kevin:NatureLover$ be the query, then \textsc{All-MinAs($Q,\cK$)} $= \{\{F_2, F_4, F_6, F_1\},$ $\{F_3, F_5, F_6, F_1\}\}$, while the pinpointing formula is
$((F_2 \wedge F_4) \vee (F_3 \wedge F_5))\wedge F_6 \wedge F_1$.
\end{small}
\end{example}

\noindent In order to build the BDDs and compute the associated probabilities, TRILL and TRILL$^P$ exploit a Prolog library of the \texttt{cplint} suite \cite{Rig09-LJIGPL-IJ}. The code of TRILL and TRILL$^P$ is available at \url{https://sites.google.com/a/unife.it/ml/trill}.

% 
% \section{LP$^2$}
% \label{LP2}
% Nowadays, we are studying a different approach to the inference on probabilistic logic programs under the distribution semantics, the same semantics of DISPONTE. 
% In the last years probabilistic logic programming received an increasing attention due to its ability to combine probability and logic programming. We are adapting the 
% Generalized Counting First Order Variable Elimination (GC-FOVE) presented in \cite{DBLP:journals/jair/TaghipourFDB13} in order to exploit lifted inference \cite{DBLP:conf/ilp/Poole08,2013arXiv1312.4328D}, 
% which generates templates, named \emph{parametric factors} or \emph{parfactors}, which stand for the actual factors found in the inference process, 
% thus delaying as much as possible the use of fully instantiated factors. We are extending the state-of-art algorithm GC-FOVE by introducing two new operators, called \emph{heterogeneous sum} 
% and \emph{heterogeneous multiplication}, in order to support reasoning on distribution semantics.
% 

\section{Related Work}
\label{related}

DL reasoners written in Prolog do not need to implement a backtracking algorithm but can exploit Prolog backtracking facilities for performing the search.
This has been observed in various works. \citeN{DBLP:journals/jar/BeckertP95} proposed a tableau reasoner in Prolog for FOL   based  on  variable-free semantic tableaux, 
but it is not tailored to DLs.
\citeN{Meissner} presented the implementation of a Prolog reasoner for the DL $\mathcal{ALCN}$. 
\citeN{Herchenroder:Thesis:2006} improved it by implementing heuristic search techniques to reduce the running time. 
\citeN{Faizi:Thesis:2011} added to its work the possibility of returning explanations for queries w.r.t. $\mathcal{ALC}$ KBs.
\citeN{DBLP:dblp_journals/iandc/HustadtMS08} presented the KAON2 algorithm that exploits basic superposition, a refutational theorem proving method for FOL 
with equality and a new inference rule, called decomposition, to reduce a $\mathcal{SHIQ}$ KB into a disjunctive datalog program.
\citeN{DBLP:dblp_journals/tplp/LukacsyS09} presented DLog, that is an ABox reasoning algorithm for the $\mathcal{SHIQ}$ language. 
It allows to store the content of the ABox externally in a database and to answer %instance check and instance retrieval 
ABox queries by transforming the KB into a Prolog program. 
TRILL and TRILL$^P$ differ from these works for the considered DL and, in particular, from DLog for the capability of answering general queries.

\citeN{DBLP:dblp_conf/ecai/BruynoogheMKGVJR10} presented FOProblog that is based on Problog, in which a program contains a set of \emph{probabilistic facts}, i.e. facts annotated with
probabilities, and a set of general clauses which can have positive and negative probabilistic facts in their body. Each fact is assumed to be probabilistically independent. 
It follows the distribution semantics and exploits BDDs to compute the probability of queries. 
FOProblog is a reasoner for FOL that is not tailored to DLs, so the algorithm could be suboptimal. It does not exploit a tableau algorithm and cannot manage probabilistic facts which are annotated with more than one probability value.

A different approach is the one of \citeN{DBLP:journals/logcom/RiccaGSDGL09} that presented OntoDLV, a system for reasoning on logic-based ontology representation language, 
called OntoDLP. OntoDLP is an extension of (disjunctive) ASP and can interoperate with OWL. OntoDLV rewrites the OWL KB into 
the OntoDLP language, can retrieve information directly from external OWL Ontologies and answers queries by using ASP.

BUNDLE \cite{RigBelLamZese13-RR13} is a probabilistic reasoner that computes the probability of queries from probabilistic KBs that follow the DISPONTE semantics. It is based 
on Pellet and is completely written in Java. It exploits a modified version of Pellet for finding the \textsc{All-MinAs} set and 
then it translates it into a BDD from which it computes the probability of the query. 
% Both TRILL and THRILL$^P$ differs from BUNDLE in the language used. They resolve different problems with respect to BUNDLE but the resulting probability values
% is the same for all the three different algorithms. 
Similarly to BUNDLE, PRONTO \cite{DBLP:conf/esws/Klinov08} is based on Pellet and performs inference on P-$\mathcal{SHIQ}$(D) KBs in which the probabilistic part contains \emph{conditional contraints} of the form $(D|C)[l,u]$ that informally 
mean ``generally, if an object belongs to $C$, then it belongs to $D$ with a probability in the interval $[l,u]$''.
P-$\mathcal{SHIQ}$(D) \cite{DBLP:journals/ai/Lukasiewicz08} uses probabilistic lexicographic entailment from probabilistic default reasoning and allows both 
terminological and assertional probabilistic knowledge about instances of concepts and roles.  
P-$\mathcal{SHIQ}$(D) is based on Nilsson's probabilistic logic \cite{DBLP:journals/ai/Nilsson86} in which the probabilistic interpretation $Pr$ defines a
probability distribution over the set of interpretations $Int$ instead of 
a probability distribution over theories. 
The probability of a logical formula $F$ according to $Pr$, denoted $Pr (F)$, is the sum of all $Pr (I)$ such that $I \in Int$ and $I \models F$.

\section{Experiments}
\label{exp}
We did several experiments in order to evaluate the performances of the algorithms we have implemented. 
Here we report a comparison between the performances of TRILL, TRILL$^P$ and BUNDLE  when computing probability for queries.
We used four different knowledge bases of various complexity:
% \begin{itemize}
%  \item BRCA\footnote{\url{http://www2.cs.man.ac.uk/~klinovp/pronto/brc/cancer_cc.owl}} models the risk factor of breast cancer;
%  \item an extract of the DBPedia\footnote{\url{http://dbpedia.org/}} ontology obtained from Wikipedia;
%  \item Biopax level 3\footnote{\url{http://www.biopax.org/}} models metabolic pathways;
%  \item Vicodi\footnote{\url{http://www.vicodi.org/}} contains information on European history.
% \end{itemize}
1) BRCA\footnote{\url{http://www2.cs.man.ac.uk/~klinovp/pronto/brc/cancer_cc.owl}} models the risk factor of breast cancer; 
2) an extract of the DBPedia\footnote{\url{http://dbpedia.org/}} ontology obtained from Wikipedia;
3) Biopax level 3\footnote{\url{http://www.biopax.org/}} models metabolic pathways;
4) Vicodi\footnote{\url{http://www.vicodi.org/}} contains information on European history.

For the tests, we used a version of the DBPedia and Biopax KBs without the ABox, a version of the BRCA with an ABox containing 1 individual and a version of Vicodi with an ABox containing 19 individuals.
To each KB, we added 50 probabilistic axioms.
For each datasets we randomly created 100 different queries. In particular, for the DBPedia and Biopax datasets we created 100 subclass-of queries while for
the other KBs we created 80 subclass-of and 20 instance-of queries.
For generating the subclass-of queries, we randomly selected two classes that are connected in the hierarchy of classes contained in the ontology, so that each query had
at least one explanation. For the instance-of queries, we randomly selected an individual $a$ and a class to which $a$ belongs by following the hierarchy of the classes 
starting from the class to which $a$ is instantiated in the KB.

Table \ref{table:res} shows, for each ontology, the average number of different MinAs computed and the average time in seconds that TRILL, TRILL$^P$ and BUNDLE took for  answering the queries.
In particular, the BRCA and the version of DBPedia that we used contain a large number of subclass axioms between complex concepts.
These preliminary tests show that both TRILL and TRILL$^P$ performances can sometimes be better than BUNDLE, even if they lack
all the optimizations that BUNDLE inherits from Pellet. This represents evidence that a Prolog implementation of a Semantic Web tableau reasoner 
is feasible and that may lead to a practical system. Moreover, TRILL$^P$ presents an improvement of the execution time with respect to TRILL 
when more MinAs are present. % while the runtime is similar to TRILL when the number of MinAs are small. 

\begin{table}
   
{
\begin{small}
\begin{tabular}[width=0.5]
%{|p{17mm}|p{5.9mm}|p{5.2mm}|p{5.2mm}|r|p{5.2mm}|} \hline 
{l|c|c|c|c} 
\textsc{Dataset} &\textsc{avg. n. minAs} &\textsc{TRILL time (s)}&\textsc{TRILL$^P$ time (s)}&\textsc{BUNDLE time (s)}\\
BRCA & 6.49 & 27.87 & 4.74 & 6.96 \\ %ELH
DBPedia & 16.32 & 51.56 & 4.67 & 3.79\\  %EL
Biopax level 3 & 3.92 & 0.12 & 0.12 & 1.85\\ %SHIN(D)
Vicodi & 1.02 & 0.19 & 0.19 & 1.12\\ %ALH(D)
\end{tabular}
\end{small}
}
 \caption{Average times for computing the probability of queries in seconds.}
    \label{table:res}

\end{table}

\section{Open Issues and expected achievement}
\label{issues-achi}
Our work aims at developing fast algorithms for performing inference over probabilistic DISPONTE semantics.
Section \ref{trill} shows that TRILL and TRILL$^P$ can compute the explanations for a query and its probability w.r.t. a 
$\mathcal{SHOIN}(\mathbf{D})$ and an $\mathcal{ALC}$ probbilistic KB respectively. For the future we plan to improve the performances of both algorithms. %study and implement optimizations for both algorithms. 
% In particular, we plan to carefully choose the rule and node application order.

We are also studying the problem of lifted inference for probabilistic logic programming using lifted variable elimination.
We are adapting the Generalized Counting First Order Variable Elimination (GC-FOVE) algorithm presented in \cite{DBLP:journals/jair/TaghipourFDB13} 
to probabilistic logic programming under the distribution semantics. To this purpose, we are developing the system LP$^2$ that extends GC-FOVE by 
introducing two new operators, \emph{heterogeneous sum} and \emph{heterogeneous multiplication}. This work will be presented at the ICLP 2014 main conference.

A second line of research is the problem of learning the parameters and the structure of a DISPONTE KB. Along this line, 
in \cite{DBLP:conf/rr/RiguzziBLZ13a} we presented a learning algorithm, called EDGE, 
that learns the parameters by taking as input a DL theory and a number of examples that
are usually concept assertions divided into positive and negative examples.
EDGE first computes, for each example, the BDD encoding its explanations, then it executes an \emph{Expectation-Maximization} (EM) algorithm, in which the functions 
Expactation and Maximization are repeatedly applied until the log-likelihood of the examples reaches a local maximum. 
Moreover, we are working on extending EDGE in order to learn also the structure of a DISPONTE KB togheter with the parameters by adapting the CELOE algorithm \cite{DBLP:dblp_journals/ws/LehmannABT11}.

\section{Conclusions}
\label{conc}
In this paper we presented two algorithms TRILL and TRILL$^P$ for reasoning on \linebreak DISPONTE KBs which are written in Prolog. 
The experiments show that Prolog is a viable language for implementing DL reasoning algorithms and
that the performances of the two presented algorithms are comparable with those of a state-of-art reasoner.

\section{Acknowledgements}
\label{ack}
This work was started by the Artificial Intelligence research group of the engineering department of the University of Ferrara.
We would personally thank my colleagues and friends (in alphabetical order) Elena Bellodi, Evelina Lamma and Fabrizio Riguzzi.

\bibliographystyle{acmtrans}

\label{lastpage}
\end{document}